\documentclass[10pt,twocolumn,letterpaper]{article}

\usepackage{cvpr}
\usepackage{times}
\usepackage{epsfig}
\usepackage{graphicx}
\usepackage{amsmath}
\usepackage{amssymb}
\usepackage{xcolor}

\usepackage[sort,nocompress]{cite}

\usepackage[pagebackref=true,breaklinks=true,letterpaper=true,colorlinks,bookmarks=false]{hyperref}

\definecolor{purple}{rgb}{0.65,0,0.65}
\newcommand{\cy}[1]{{\color{black}#1}}

\newcommand{\rz}[1]{{\color{black}#1}}
\newcommand{\kx}[1]{{\color{black}#1}}

\cvprfinalcopy % *** Uncomment this line for the final submission

\def\cvprPaperID{6607} % *** Enter the ICCV Paper ID here

\definecolor{purple}{rgb}{0.65,0,0.65}
\definecolor{turquoise}{rgb}{0.0,0.5,0.5}
\newcommand{\hao}[1]{{\color{black}#1}}

\ifcvprfinal\fi
\begin{document}

	%%%%%%%%% TITLE
	%\title{CoSegNet: Co-segmentation of 3D Shape Sets with Online Learned Networks}
	%\title{CoSegNet: Deep Co-Segmentation of 3D Shapes via Online Learning}
%	\title{CoSegNet: Deep Co-Segmentation of 3D Shapes with Group Consistency Loss}
	\title{AdaCoSeg: Adaptive Shape Co-Segmentation with Group Consistency Loss}

	\author{
		Chenyang Zhu$^{1,2}$ \quad
		Kai Xu$^{2}$\thanks{Corresponding author: kevin.kai.xu@gmail.com} \quad
		Siddhartha Chaudhuri$^{3,4}$ \quad
		Li Yi$^{5}$ \quad
		Leonidas Guibas$^{6}$ \quad
		Hao Zhang$^{1}$\\
		$^1$Simon Fraser University \quad $^2$National University of Defense Technology\\
		$^3$Adobe Research \quad $^4$IIT Bombay \quad $^5$Google Research \quad $^6$Stanford University
	}

	\maketitle

	\begin{abstract}
		We introduce AdaCoSeg, a deep neural network architecture for {\em adaptive co-segmentation\/} of a set of 3D shapes represented as point clouds.
		\hao{Differently from the familiar single-instance segmentation problem, co-segmentation is intrinsically {\em contextual\/}: how a
			shape is segmented can vary depending on the set it is in. Hence, our network features an adaptive learning module to produce
			a consistent shape segmentation which adapts to a set. Specifically, given an input set of unsegmented
			shapes, we first employ an offline pre-trained {\em part prior network\/} to propose per-shape parts. Then, the co-segmentation
			network iteratively and} jointly optimizes the part labelings across the set subjected to a novel {\em group consistency loss\/} defined
		by matrix ranks. \hao{While the part prior network can be trained with noisy and inconsistently segmented shapes, the final
			output of AdaCoSeg}
		%
		%The part proposals are refined in each iteration by an auxiliary network that acts as a weak regularizing prior, pre-trained to denoise noisy,
		% unlabeled parts from a large collection of segmented 3D shapes, where the part compositions within the same object category can be
		%highly inconsistent. The output
		%
		is a consistent part labeling for the input set, with each shape segmented into up to (a user-specified) $K$ parts.
		Overall, our method is {\em weakly supervised}, producing segmentations tailored to the test set, without
		consistent ground-truth segmentations. We show qualitative and quantitative results from AdaCoSeg and evaluate it via ablation studies
		and comparisons to state-of-the-art co-segmentation methods.
	\end{abstract}

	% -*- compile-command: "texify --pdf --quiet decop_cvpr18.tex" -*-
% !TEX root = decop_cvpr18.tex
% (shell-command "start decop_cvpr18.pdf")

\section{Introduction}
\label{sec:intro}

With the proliferation of data-driven and deep learning techniques in computer vision and computer graphics, remarkable progress has been made on
supervised image~\cite{DBLP:journals/corr/BadrinarayananK15,DBLP:journals/corr/ChenPK0Y16}
%~\cite{lin2014,long2015,DBLP:journals/corr/BadrinarayananK15,DBLP:journals/corr/ChenPK0Y16,DBLP:journals/corr/LinMS016,DBLP:journals/corr/PengZYLS17}
and shape segmentations~\cite{kalogerakis20173d,yi2017syncspeccnn}.
%~\cite{qi2017pnpp,riegler2017octnet,wang2017cnn,kalogerakis20173d,yi2017syncspeccnn}.
%
{\em Co-segmentation\/} is an instance of the segmentation problem where the input consists of a {\em collection\/}, rather than one piece, of data
and the collection shares certain common characteristics. Typically, for shape co-segmentation, the commonality is that the shapes all belong to the same category, e.g., chairs or airplanes.

\begin{figure}[t!]
	\includegraphics[width=\linewidth]{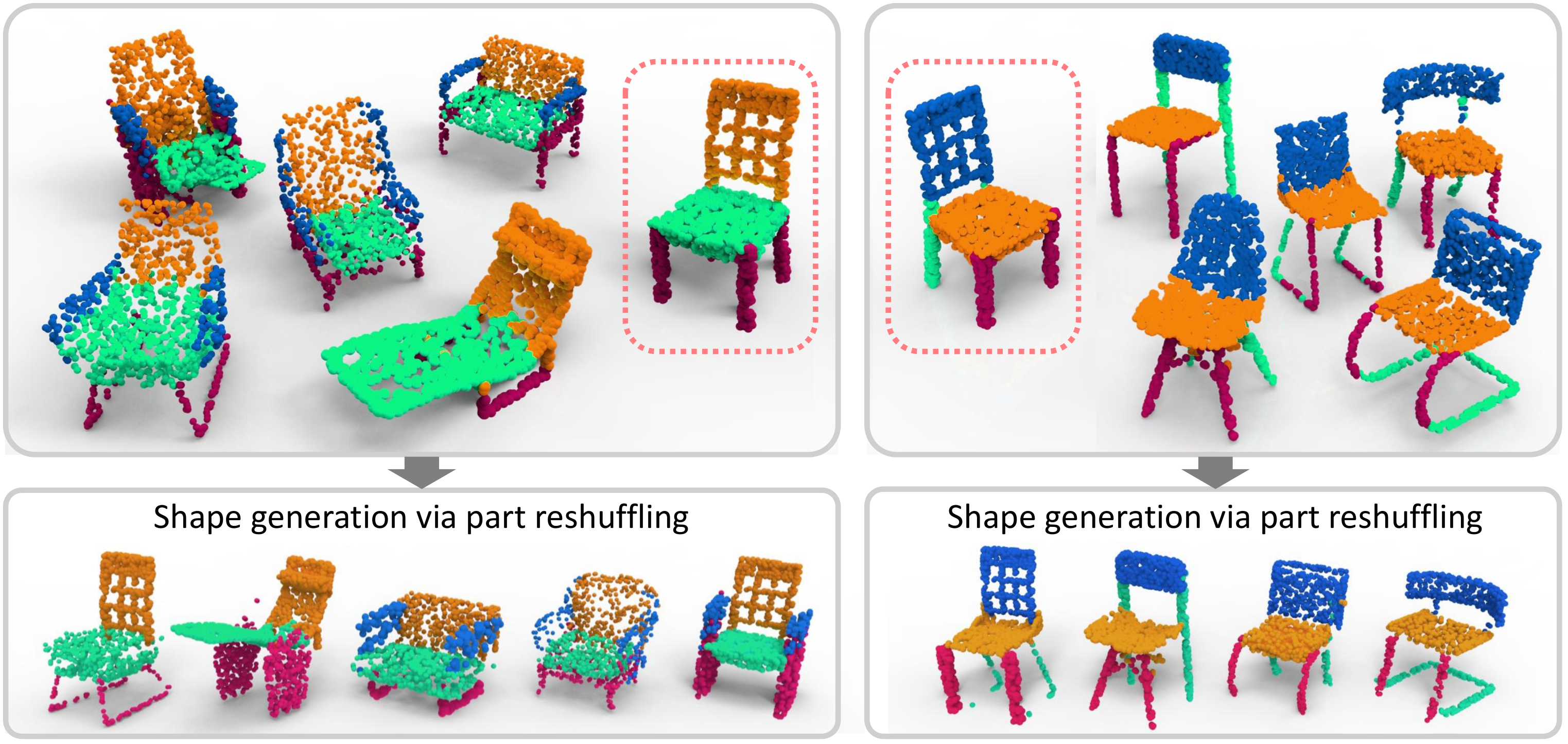}
	\caption{Our adaptive shape co-segmentation network, \mbox{AdaCoSeg}, produces structurally different segmentations (here up to $4$ parts) for two sets of chairs --- one with armrests, one without. For each set, the segmentations are semantically consistent, allowing shape generation via part reshuffling. However, the same shape can be segmented differently depending on its containing set (see the circled chair), showing the method's adaptivity.}
	\label{fig:teaser}
\end{figure}

The goal of co-segmentation is to compute a {\em consistent\/} segmentation for all shapes in the input collection.
The consistency of the segmentation implies a correspondence between all the segmented parts, which is a critical requirement for knowledge and attribute transfer, collecting statistics over a dataset, and structure-aware shape modeling~\cite{mitra2013structure}. Figure~\ref{fig:teaser} shows such a modeling example based on part reshuffling induced by a co-segmentation.

In contrast to the familiar single-instance segmentation problem, a distinctive feature of co-segmentation is that it is inherently \emph{contextual}.
As dictated by the consistency criterion, the same shape may be segmented differently depending on which input set it belongs to; see Figure~\ref{fig:teaser}.
From this perspective, the input shape collection serves both as the test set {\em and\/} the training set. Ideally, the
co-segmentation network can {\em quickly adapt\/} to a new input set without expensive retraining. Such an adaptive network would change its
behavior, i.e., the network weights, at the time it is run. This is different from the traditional label learning paradigm, where the trained model strives
to generalize to new inputs without changing the network weights, either under the supervised~\cite{qi2017pnpp,kalogerakis20173d} or weakly supervised settings~\cite{sung2018deep,muralikrishnan2018tags2parts,chen2019baenet}.

%
%This is where a connection can be made between
%co-segmentation and {\em online learning\/}~\cite{shalev2007online,sahoo2017online}, beyond the traditional offline label learning paradigms~\cite{sidi2011unsupervised,hu2012co,sung2018deep,muralikrishnan2018tags2parts}.
%
%: A co-segmentation method should contain an \emph{online} stage which is able to quickly adapt to new input sets without expensive retraining.
%On the other hand, most existing approaches to shape co-segmentation solve the learning problem directly
%over the input set~\cite{sidi2011unsupervised,hu2012co} where the model is neither adaptive and nor scalable.

%\rz{Most shape co-segmentation methods to date have been unsupervised or weakly supervised~\cite{golovinskiy2009consistent,sidi2011unsupervised,hu2012co,wang2012active,vanKaick2013}, and this is due to two main reasons. First, accurate and consistent ground-truth segmentations are difficult to obtain, especially for large collections. An examination of existing large repositories, e.g.,~\cite{sung2017comp,yi2016scalable}, reveals that the shape segmentations therein can be highly inconsistent (e.g., see Figure~\ref{fig:inconsistentpart}).
%Second, as dictated by the consistency criterion, the same shape may be segmented differently depending on which collection it belongs to (Figure~\ref{fig:teaser}). In a way, the input shape collection serves as both training set and test set. Ideally, the learning scheme for co-segmentation should avoid expensive retraining and quickly adapt to new input sets.}

In this paper, we introduce a deep neural network for shape co-segmentation, coined AdaCoSeg, \hao{which is designed to be adaptive.}
% Our network takes as input 3D shapes represented by point clouds.
AdaCoSeg takes as input a set of unsegmented shapes represented as point clouds, proposes {\em per-shape\/} parts in the first stage, and then {\em jointly optimizes} the parts subject to a novel {\em group consistency loss\/} defined by {\em matrix rank estimates\/} \hao{for the specific input set.} The output is a $K$-way consistent part labeling for each shape, where $K$ is a user-specified hyperparameter for the network. The network weights are initialized randomly and iteratively optimized via backpropagation based on the group loss.

While the co-segmentation component is unsupervised, guided by the group consistency loss, we found that the results can be improved by adding a weak regularizing prior to \hao{boost the part proposal.}
\kx{Specifically, we pre-train a {\em part prior network} which takes as input a possibly noisy proposed part, represented by an indicator function over the complete point cloud, and denoises or ``snaps'' it to a more plausible and clean part.} The part prior network is similar to the pairwise potential of a conditional random field (CRF) in traditional segmentation~\cite{kalogerakis2010crf}: while it is not a general prior, as it is trained to remove only a small amount of noise, it suffices for boundary optimization. It is trained on a large collection of segmented 3D shapes, e.g., ShapeNet~\cite{chang2015shapenet}, where part counts and part compositions within the same object category can be highly inconsistent. No segment label is necessary: the model is label-agnostic.

\rz{Overall, our method is {\em weakly supervised\/}, since it produces consistent segmentations without consistent ground-truth segmentations.}
It consists of an offline, \rz{supervised} part prior network, \rz{which is trained once on inconsistently segmented, unlabeled shapes}, and a ``runtime'', \hao{adaptive} co-segmentation network which is unsupervised and executed for each input set of shapes. It is important to note that consistency of the segmentations is not tied to the part count $K$, but to the geometric and structural features of the shape parts in the set, with $K$ serving as an {\em upper bound\/} for the part counts; see Figure~\ref{fig:teaser}. On the other hand, adjusting $K$ allows AdaCoSeg to produce consistent co-segmentations at varying levels of granularity; see Figure~\ref{fig:hierarchy}.

\begin{figure}[t!]
	\includegraphics[width=\linewidth]{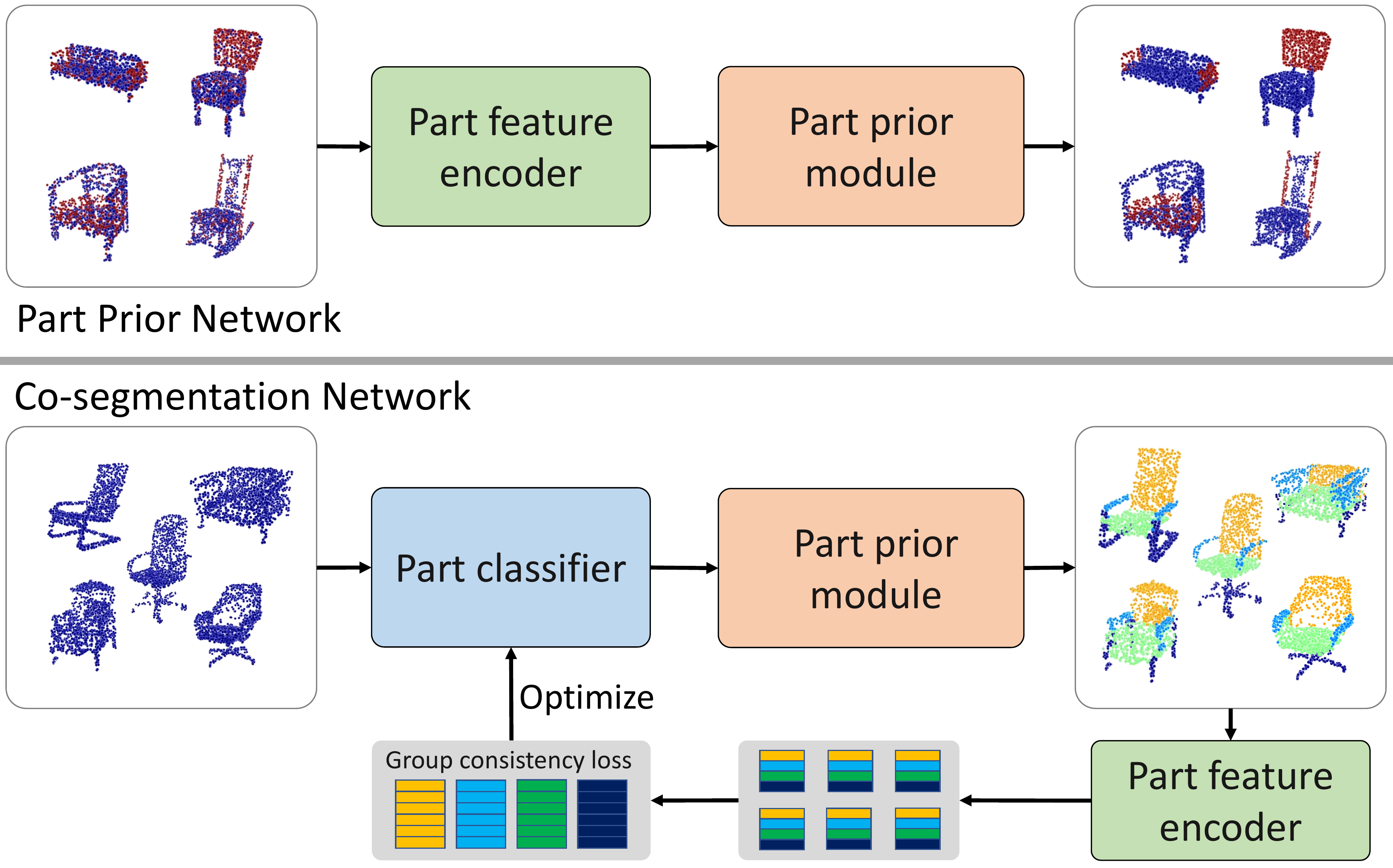}
	\caption{AdaCoSeg consists of a part prior network (top) and a co-segmentation network (bottom). The part feature encoder and part prior module in the first network learn a weak regularizing prior to denoise proposed part shapes. The co-segmentation network is trained with a novel group consistency loss, defined on a set of shapes, based on the ranks of part similarity matrices.}
	\label{fig:overview}
\end{figure}

Our part prior network is trained using the dataset from ComplementMe~\cite{sung2017comp}; the
adaptive co-segmentation is unsupervised. For evaluation only, we also adopt two datasets~\cite{wang2012active,yi2016scalable} containing ground truth co-segmentations.
While offline training required up to $20$ hours to complete, it takes about \kx{$7$ minutes} to co-segment $20$ shapes at a resolution of 2,048 points per shape.
We show qualitative and quantitative results from AdaCoSeg and evaluate it through ablation studies and
comparisons with state-of-the-art co-segmentation methods. Our main contributions include:

\begin{itemize}
	\vspace{-6pt}
	\item The first DNN for adaptive shape co-segmentation.
	\vspace{-6pt}
	\item A novel and effective group consistency loss based on low-rank approximations.
	\vspace{-6pt}
	\item A co-segmentation training framework that needs no ground-truth consistent segmentation labels.
\end{itemize}

\section{Related work}
\label{sec:related}

\paragraph{Deep learning for shape segmentation.}
Deep models for supervised shape segmentation have been developed for various representations, such as voxel grids~\cite{riegler2017octnet,wang2017cnn}, point clouds~\cite{qi2017pnpp,klokov2017escape,huang2018recurrent}, multi-view projections~\cite{kalogerakis20173d}, and surface meshes~\cite{yi2017syncspeccnn,wang20183d}. 
The key is to replace hand-crafted features employed in traditional methods by features learned from data. However, these models are mostly trained to target a \emph{fixed} set 
of semantic labels. \hao{The resulting segmentation for a given shape is also fixed and cannot be adaptive to the
context of a shape set,} a key feature of co-segmentation.
%They cannot determine the label set dynamically based on the shapes being segmented, which is .
%Moreover, it is costly to obtain a large training dataset with different target label sets.
Relatively few works study deep learning for unsupervised shape segmentation~\cite{shu2016unsup,chen2019baenet}.
%where unsupervised learning is used only for feature learning but not for segmentation itself.

\vspace{-10pt}

\paragraph{Image co-segmentation.}
The co-segmentation of a pair or a group of 2D images has been studied for many years in the field of computer vision, where the main goal is to segment out a common object from multiple images~\cite{vicente2011object}. Most works formulate this problem as a multi-image Markov Random Field (MRF), with a foreground consistency constraint. 
%Foreground consistency is measured based on color or object shape similarity, or dense image correspondence~\cite{wang2013image}. 
Recently, Li et al.~\cite{li2018deep} proposed a deep Siamese network to achieve object co-extraction from a pair of images. The general problem setting for all of these image co-segmentation works is significantly different from ours.

\vspace{-10pt}

\paragraph{Shape co-segmentation.}
%
%Since the seminal work of consistent segmentation of 3D shapes by~\cite{golovinskiy2009consistent}, 
Extensive research has been devoted to co-analysis of sets of shapes~\cite{golovinskiy2009consistent,xu2010style,sidi2011unsupervised,huang2011joint,hu2012co,wang2012active}. These methods often start with an over-segmentation and perform feature embedding and clustering of the over-segmented patches to obtain a consistent segmentation. While most of these methods are unsupervised, their analysis pipelines all adopt hand-craft features \rz{and heuristic-based clustering, often leading to unnatural results amid complex part or structure variations.} 
%In contrast, our part prior network learns the part features from a large dataset and our co-segmentation network learns the network weights through a joint optimization.

Recently, deep learning based approaches are emerging. Shu et al.~\cite{shu2016unsup} use deep auto-encoders for per-part feature learning. \cy{However, their co-segmentation module does not use a deep network and it strictly constrains the final segmentations to parts learned in the first stage. In contrast, AdaCoSeg does not strictly adhere to proposals by the part prior network, as the consistency loss can impact and adjust part labeling.}
Muralikrishnan et al.~\cite{muralikrishnan2018tags2parts} propose a weakly-supervised method for tag-driven 3D shape co-segmentation, but their model is trained to target a pre-defined label set.
%We are not aware of a deep learning model developed for co-segmentation of 3D shape sets. This may be due to the fact that incremental learning remains under-developed for DNNs~\cite{sahoo2017online}.
Sung et al.~\cite{sung2018deep} attempt to relate a set of shapes with deep functional dictionaries, resulting in a co-segmentation. However, these dictionaries are learned offline, 
for individual shapes, %, and there is not an online segmentation process.
so their model cannot \hao{adaptively} co-segment a set of shapes. In contrast, CoSetNet is split into an offline part which is transferrable
across different shape sets, and an online, adaptive co-segmentation network which is learned for a specific input set.

\hao{In concurrent work, Chen et al.~\cite{chen2019baenet} present a branched autoencoder for weakly supervised shape co-segmentation. The key difference is that BAE-NET is essentially a more advanced part prior network, with each branch tasked to learn a simple representation for one universal part of an input shape collection; there is no explicit optimization for group consistency. As a result, BAE-NET tends to under-perform comapred to AdaCoSeg on small input sets and in the presence of large part discrepancies; see Figure~\ref{fig:comp-bae}.}
%

%\paragraph{Online learning.}
%%
%Online learning represents a family of machine learning algorithms that learn to update models incrementally from sequentially input data streams~\cite{shalev2007online,jin2010online}.
%In recent years, online Learning has be directly applied to DNNs through online
%backpropagation. However, such direct method suffers from issues like vanishing gradient and diminishing feature reuse. More non-trivial attempts for online deep learning typically adopt a sliding window scheme in mini-batch training~\cite{zhou2012online,lee2016dual}, which is unsuitable for a streaming data setting.
%Recently, Sahoo et al.~\cite{sahoo2017online} propose am online deep learning framework which learns
%DNN models of adaptive depth from a sequence of training data in an online learning setting.
%In our setting, the segmentation model should adapt its output layer (targeted labels), instead of network depth, to different input shape sets. This, to our knowledge, has not been studied previously. 

	% -*- compile-command: "texify --pdf --quiet decop_cvpr18.tex" -*-
% !TEX root = decop_cvpr18.tex
% (shell-command "start decop_cvpr18.pdf")

\section{Overview}
\label{sec:overview}

Our method works with point-set 3D shapes and formulates shape segmentation as a
point labeling problem. The network has a two-stage architecture; see Figure~\ref{fig:overview}.

% contains two stages: an offline training of a \emph{part labeling
%refinement network} and a runtime joint optimization by a novel \emph{co-segmentation network}.

\vspace{-10pt}

\paragraph{Part prior network.}
The network takes as input a point cloud with noisy binary labeling, where the foreground
represents an imperfect part, and outputs a regularized labeling leading to a refined part.
To train the network, we employ the ComplementMe dataset~\cite{sung2017comp}, a subset of
ShapeNet~\cite{chang2015shapenet}, which provides semantic part segmentation.
The 3D shapes are point sampled, with each shape part implying a binary labeling.
For each binary labeling, some random noise is added; the part prior network is trained to denoise
these binary labelings. Essentially, the part prior network learns what a valid part looks like through training on a labeling denoising task. Meanwhile, it also learns a multi-scale and part-aware shape feature at each point, which can be used later in the co-segmentation network.

\vspace{-10pt}

\paragraph{Co-segmentation network.}
Given an input set of 3D shapes represented by point clouds, our co-segmentation network learns the optimal network weights through backpropagation based on a group consistency loss defined over the input set. The network outputs a $K$-way labeling for each shape, with semantic consistency, where $K$ is a user prescribed network parameter specifying an upper bound of part counts; the final part counts are determined based on the input shape set and network optimization.

The co-segmentation network is unsupervised, without any ground-truth consistent segmentations.
For each part generated by the $K$-way classification, a binary segmentation is formed and
fed into the pre-trained part prior network: (1) to compute a refined $K$-part segmentation,
and (2) to extract a part-aware feature for each point. These together form a part feature for each segment.
The corresponding part features with the same label for all shapes in the set constitute a \emph{part feature matrix}. Then, weights of the co-segmentation network are optimized with the objective to maximize
the part feature similarity within one label and minimize the similarity across different labels.
This amounts to minimizing the rank of the part feature matrix for each semantic label
while maximizing the rank of the joint part feature matrix for two semantic labels.

	\section{Method}
\label{sec:method}

The offline stage of AdaCoSeg learns a weak \rz{regularizing} prior for plausible shape parts, where a part prior network %for (unlabeled) parts.
is trained on a large, diverse shape repository with generally inconsistent, unlabeled segmentations. The network serves to refine any proposed parts to better resemble observed ones. The runtime stage jointly analyzes a set of test shapes using a co-segmentation network that iteratively proposes (at most) $K$-way segmentations of each shape to optimize a group consistency score over the test set.

%\subsection{Network architecture}
%Our network is shown in Fig.~\ref{fig:overview}, which is composed of two modules:

\subsection{Part Prior Network}
\label{sec:offlineNet}

\paragraph{Dataset.}
In offline pre-training, we want to learn a general model \rz{to denoise} all plausible part shapes at all granularities, using off-the-shelf data available in large quantities. This weak prior will be used to regularize any consistent segmentation of test shapes. Repositories with standard labeled segmentations~\cite{wang2012active,yi2016scalable} are both limited in size and fixed at single pre-decided granularities. Instead, we use the 3D part dataset developed for ComplementMe~\cite{sung2017comp}.

This dataset, a subset of ShapeNet~\cite{chang2015shapenet}, exploits the fact that shapes in existing 3D repositories already have basic component structure, since artists designed them modularly. However, the segmentations are inconsistent: while a chair back may be an isolated part in one shape, the back and seat may be combined into a single part in another. ComplementMe does some basic heuristic-based merging of adjacent parts to eliminate very small parts from the collection, but otherwise leaves noisy part structures untouched. Further, the parts lack labels -- while some tags may be present in the input shapes, we ignore them since the text is generally inconsistent and often semantically meaningless. Hence, this dataset is an excellent example of the weakly-supervised training data we can expect in a real-life situation. Our method trains a \rz{denoising} prior on this noisy dataset, which will be used to refine consistent segmentations proposed in our co-segmentation stage.
%\sid{Need to mention that we do category-specific offline training.}

\begin{figure}[t!]
	\includegraphics[width=\linewidth]{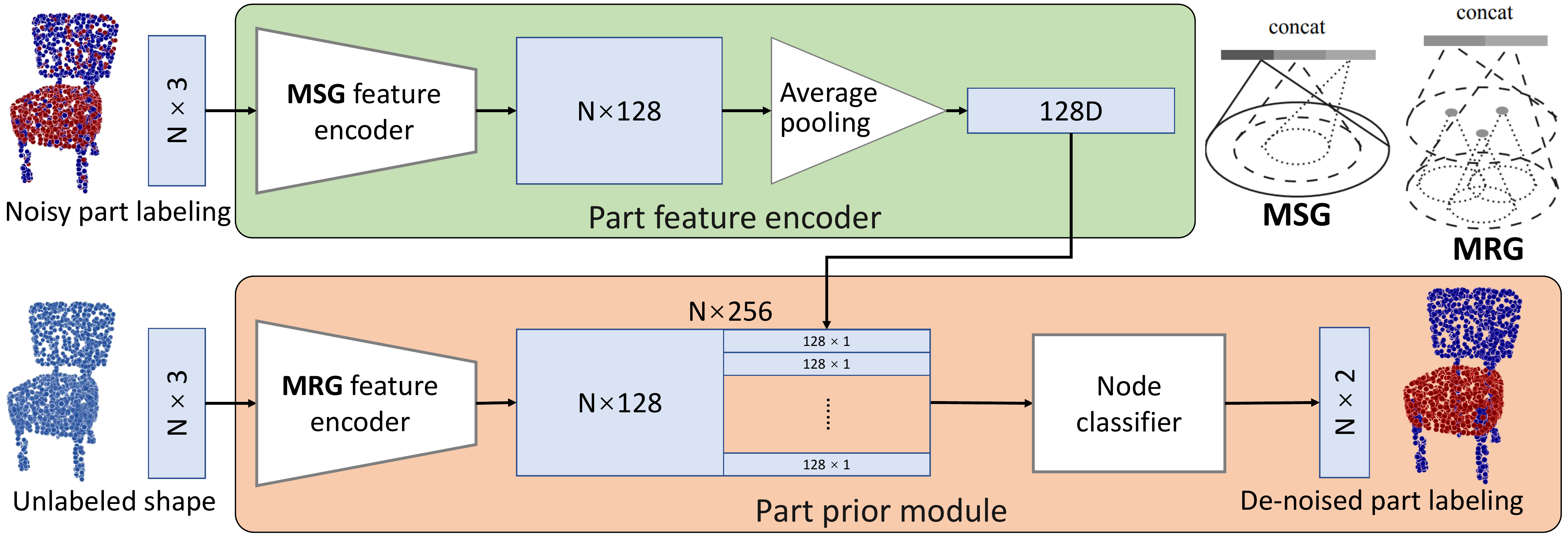}
	\caption{The architecture of the part prior network. The network encodes a shape with noisy part labeling and the whole shape, using the MSG and MRG feature encoders from PointNet++~\cite{qi2017pnpp}, respectively. It is trained to denoise the input binary labeling and output a clean labeling, indicating a plausible part.}
	\label{fig:offlineNet}
\end{figure}

\begin{figure*}[t!]
	\centering
	\includegraphics[width=\linewidth]{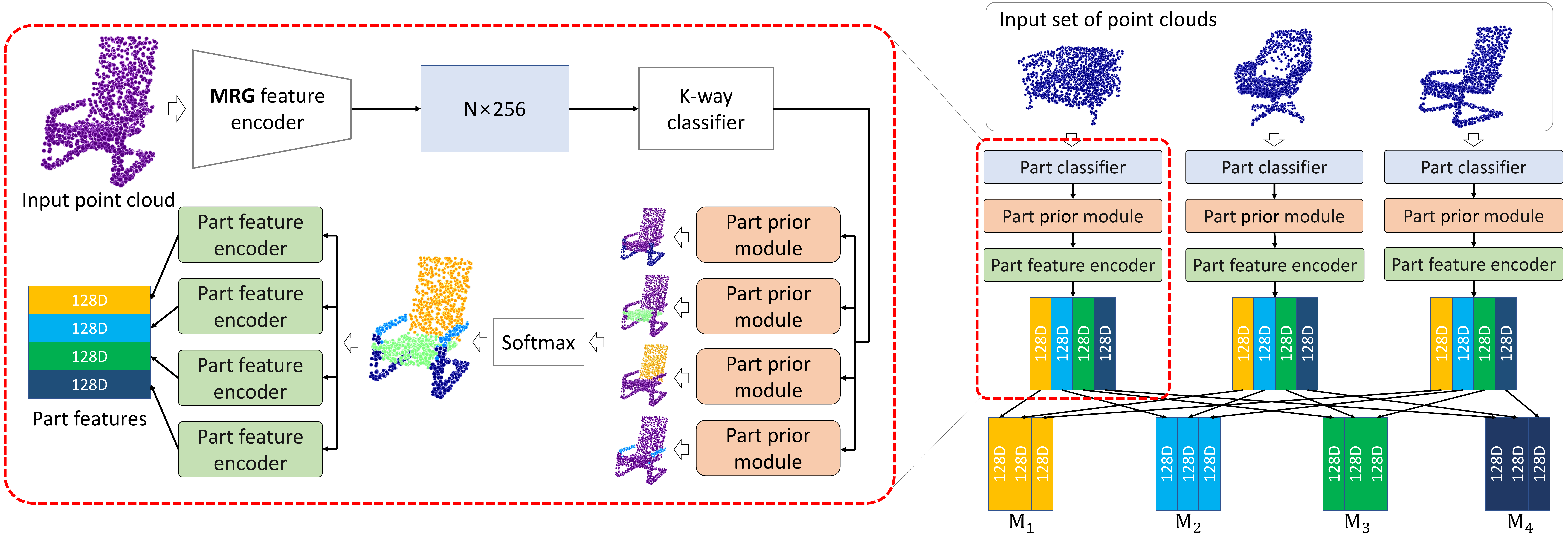}
	\caption{Left: Given an input point cloud, the $K$-way classifier segments it into $K$ parts. These parts are then refined by the part prior module, resulting in a refined $K$-way segmentation of the input point cloud. After that, the part feature encoder is used to extract features for each refined part. Right: Given a set of input point clouds, we construct a part similarity matrix for each abstract part label, based on the part features extracted for all shapes.}
	\label{fig:onlineNet}
\end{figure*}

\vspace{-8pt}

\paragraph{Network architecture.}
The part prior network learns to denoise an imperfectly segmented part, using %, e.g., by refining its boundaries.
an architecture based on components from PointNet++~\cite{qi2017pnpp}. The input to the network is a 3D point cloud shape $S$. Points belonging to the proposed part constitute the foreground $F \subset S$, while the remaining points are the background $B = S \setminus F$. The output of the network is a probability for each point $q \in S$, such that the high probability points collectively define the ideal, ``clean'' part that best matches the proposed part, thereby denoising the noisy foreground.

The architecture of our network is shown in Figure \ref{fig:offlineNet}. The point cloud is processed by the multi-scale grouping (MSG) and multi-resolution grouping (MRG) modules of PointNet++, to produce two context-sensitive 128-D feature vectors $f_\text{MSG}(q)$ and $f_\text{MRG}(q)$ for each point $q \in S$. The MSG module captures the context of a point at multiple scales, by concatenating features over larger and larger neighborhoods. The MRG module computes a similar multi-scale feature, but (half of) the features of a large neighborhood are computed recursively, from the features of the next smaller neighborhood; see ~\cite{qi2017pnpp} for details. %\sid{Why the choice of MSG in one branch and MRG in the other?}

We average the MSG features of foreground points to obtain a robust descriptor $f_\text{fg}$, which is concatenated with the MRG feature of each point to produce \mbox{$[f_\text{MRG}(q), f_\text{fg}]$} pairs. The pairs are fed to a binary classifier with ReLU activation, where the output of the classifier indicates the ``cleaned'' foreground and background.

\vspace{-8pt}

\paragraph{Training.} The part prior network is trained with single parts from the inconsistently segmented dataset. We add noise to each part (foreground) by randomly inserting some background points and excluding some foreground points \rz{($\sim$20-30\%)}.
%to simulate imperfect segmentations.
%\kx{This simulates the real noise pattern resulted from the binary segmentation of the part classifier: over- or under-segmentation of a part would mistakenly include or exclude background points.}
The network takes noisy parts as input and tries to output clean part indicator functions, using a negative log-likelihood loss and Adam~\cite{kingma2015adam} optimizer.

\subsection{Co-segmentation Network}
\label{sec:onlineNet}

The runtime stage of our pipeline jointly segments a set of unsegmented test shapes \mbox{$T = \{ S_1, S_2, \dots,  S_N \}$} to maximize consistency between the segmented parts. To this end, we design a deep neural network that takes a shape's point cloud as input and outputs a $K$-way segmentation; $K$ is a user-specified hyperparameter specifying the part count. These outputs are compared across the test set to ensure geometric consistency of corresponding segments: our quantitative metric for this is a {\em group consistency energy}, which is used as a loss function to iteratively refine the output of the network using back-propagation.

Note that although we use a deep network to output per-shape segmentation maps, the trained network is not expected to generalize to new shape sets. Hence, the network performs essentially an unsupervised $K$-way clustering of the input points across all test shapes. Apart from the consistency loss, the network is guided by the offline prior that has \rz{learned to denoise} plausible parts of various sizes, but has no notion of consistency or desired granularity.

\vspace{-8pt}

\paragraph{Network architecture.} Our co-segmentation architecture is shown in Figure \ref{fig:onlineNet}. The network takes a minibatch of test shapes as input. The first part of the network is a classifier that independently assigns one of $K$ abstract labels \mbox{$\{ L_1, L_2, \dots, L_K \}$} to each point in each shape, with shared weights: the set of points in a shape with label $L_i$ defines a single part with that label. Since the classifier output may be noisy, we pass the binary foreground/background map corresponding to each such part through the pre-trained (and frozen) offline denoising network (Section \ref{sec:offlineNet}) and then re-compose these maps into a $K$-way map using a $K$-way softmax at each point to resolve overlaps. The recomposed output is the final (eventually consistent) segmentation.

%\begin{figure}[t!]
%	\includegraphics[width=\linewidth]{figures/coseg_net.pdf}
%	\caption{Given a set of input point clouds, we construct a part similarity matrix for each abstract part label, based on the part features extracted for all shapes.}
%	\label{fig:cosegnet}
%  \vspace{-3mm}
%\end{figure}

The subsequent stages of the network are deterministic and have no trainable parameters: they are used to compute the group consistency energy. First, the MSG features~\cite{qi2017pnpp} of the foreground points for each part are max-pooled to yield a part descriptor (we found max pooling to work better than average pooling). If the segmentation is consistent across shapes, all parts with a given label $L_i$ should have similar descriptors. Therefore, we stack the descriptors for all parts with this label from all shapes in a matrix $M_i$, one per row, and try to minimize its second singular value, a proxy for its rank (low rank $=$ more consistent). Also, parts with different labels should be distinct, so the union of the rows of matrices $M_i$ and $M_{j \neq i}$ should have high rank. This time, we want to {\em maximize} the second singular value of $\mathit{concat}(M_i, M_j)$, where the $\mathit{concat}$ function constructs a new matrix with the union of the rows of its inputs. The overall energy function is:

\[
\begin{aligned}
\mathcal{E}_\text{coseg} = 1 & + \max_{i \in \{1,2,\dots,K\}} \mathit{rank}(M_i) \\
& - \min_{i,j \in \{1,2,\dots,K\}, i \neq j} \mathit{rank} \left( \mathit{concat}(M_i, M_j) \right),
\end{aligned}
\]

where the $\mathit{rank}$ function is the second singular value, computed by a (rather expensive) SVD decomposition~\cite{yi2018faces}. As this energy is optimized by gradient descent, the initial layers of the network learn to propose more and more consistent segmentations across the test dataset. Additionally, we found that gaps between segments of a shape appeared frequently and noticeably before re-composition, and were resolved arbitrarily with the subsequent softmax. Hence, we add a second energy term that penalizes such gaps; see more details in the supplementary material.

Because \rz{the co-segmentation network has no access to ground truth and relies only on a weak geometry denoising prior}, the consistency energy is the principal high-level influence on the final segmentation. We experimented with different ways to define this energy, and settled on SVD-based rank approximation as the best one. Note that the SVD operation makes this a technically non-decomposable loss, which usually needs special care to optimize~\cite{kar2014nondecomp}. However, consistency is in general a transitive property (even though its converse, inconsistency, is not). Hence, enforcing consistency over each of several overlapping batches is sufficient to ensure consistency over their union, and we can refine the segmentation maps iteratively using standard stochastic gradient descent.

	% -*- compile-command: "texify --pdf --quiet decop_cvpr18.tex" -*-
% !TEX root = decop_cvpr18.tex
% (shell-command "start decop_cvpr18.pdf")

\section{Results and Evaluations}
\label{sec:result}

We validate the two stages of AdaCoSeg through qualitative and quantitative evaluation, and compare to state-of-the-art methods. We train our part prior network on the shape part dataset from ComplementMe~\cite{sung2017comp}, which is a subset of ShapeNet~\cite{chang2015shapenet}, and test our method with the ShapeNet~\cite{yi2016scalable} and COSEG~\cite{wang2012active} semantic part datasets. We also manually labeled some small groups (6-12 shapes per group) of shapes from ShapeNet~\cite{yi2016scalable} to form a co-segmentation benchmark for quantitative evaluation.

\begin{figure}[!t]
	\includegraphics[width=\linewidth]{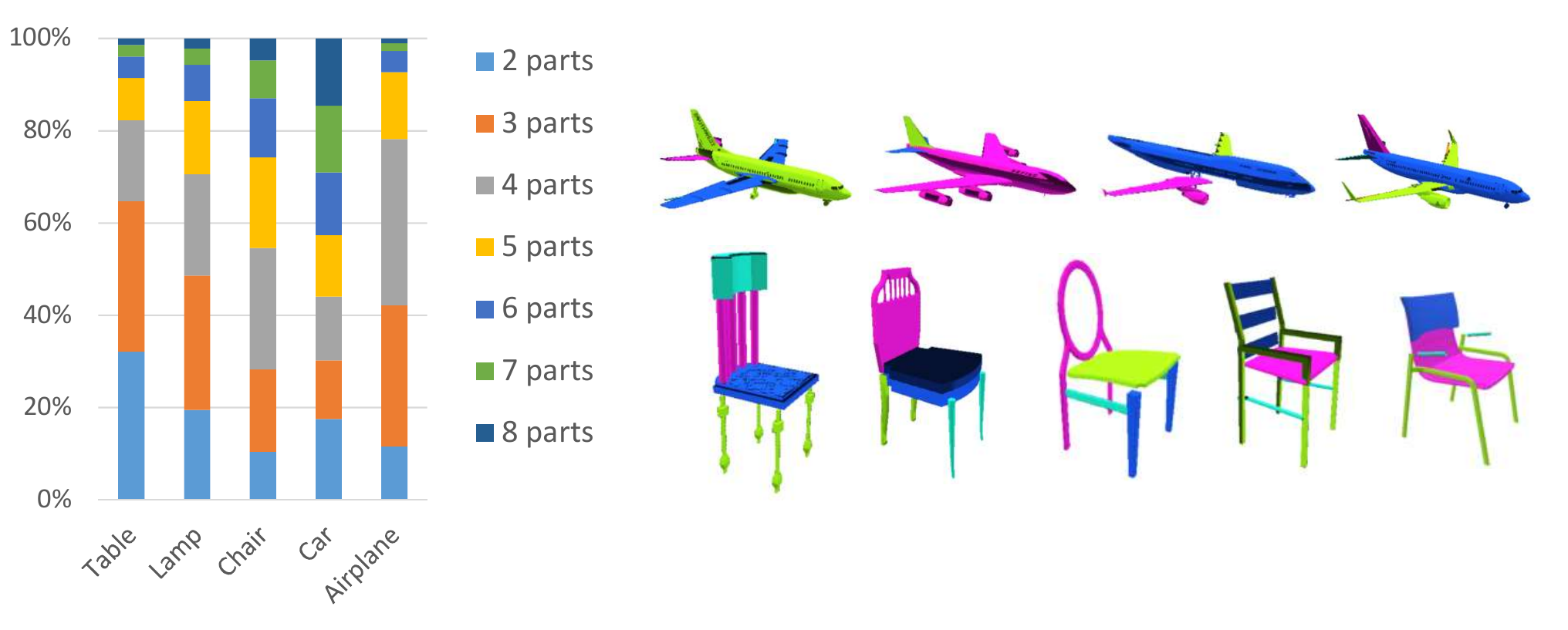}
	\caption{High degrees of inconsistencies exist in the shape segmentations available in the ComplementMe dataset~\cite{sung2017comp}. The left figure charts the distribution of part counts in each object category, showing their diversity. The right figure shows several shapes, within the same category and having the same part counts (3 parts for airplanes, 4 parts for chairs), that exhibit much structural and geometric variation in their segmentations.}
	\label{fig:inconsistentpart}
\end{figure}

\begin{table}[!t]
	\caption{Dataset for training the part prior network. For each category, we list the shape count (\#S) and part count (\#P).}
	\centering
%	\begin{tabular}{l|r|r}
%	       {\bf Category} & {\bf \# Shapes} & {\bf \# parts} \\
%		\hline
%		\hline
%		Airplane & $2,410$ & $9,134$ \\
%		Bicycle & $49$ & $299$ \\
%		Car & $976$ & $5,119$ \\
%		Chair & $2,096$ & $9,433$ \\
%		Lamp & $862$ & $3,296$ \\
%		Table & $1,976$ & $6,608$
%	\end{tabular}
        \vspace{3pt}
	\begin{tabular}{l|r|r|r|r|r|r}
	        & {\small Airplane} & {\small Bicycle} & {\small Car} & {\small Chair} & {\small Lamp} & {\small Table} \\
		\hline
		\hline
		{\small \#S} & {\small 2,410} & {\small 49} & {\small 976} & {\small 2,096} & {\small 862} & {\small 1,976} \\
               {\small \#P} & {\small 9,134} & {\small 299} & {\small 5,119} & {\small 9,433} & {\small 3,296} & {\small 6,608}
	\end{tabular}
	\label{tab:dataset}
\end{table}

\vspace{-8pt}

\begin{figure}[t!]
	\includegraphics[width=\linewidth]{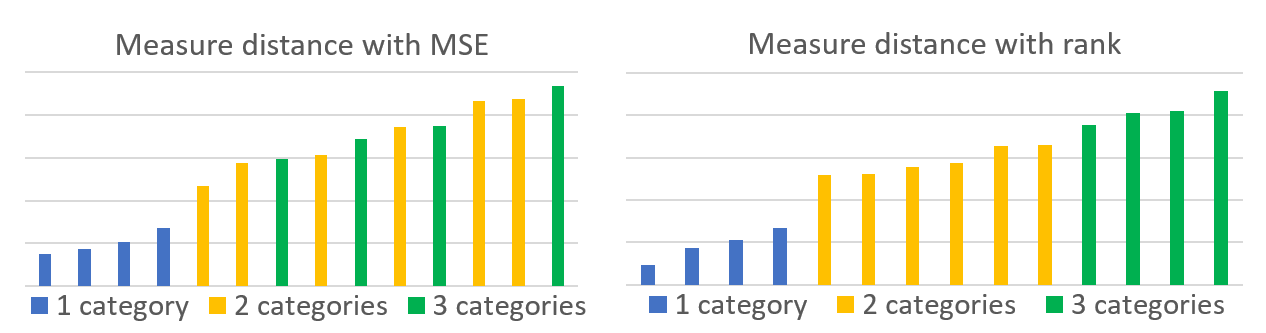}
	\caption{Number of distinct labels in a collection of parts (Y axis) vs increasing feature variation for that collection (X axis). The plot on the right uses the more discriminative matrix rank-based score, whereas the plot on the left uses MSE which cannot tell 2 and 3-label collections apart.}
	\label{fig:partFeatureDistance}
\end{figure}

\paragraph{Discriminative power of matrix ranks.}
Our network design makes a low-rank assumption for the features of corresponding shape parts: the MSG feature vectors of similar parts form a low-rank matrix, while those dissimilar parts form a higher-rank matrix, where rank is estimated in a continuous way as the magnitude of the second singular value. To show that matrix ranks provide a discriminative metric, we use the ShapeNet semantic part dataset~\cite{yi2016scalable}, which has a consistent label for each part, as test data. The chair category for this dataset has four labels: \emph{back}, \emph{seat}, \emph{arm} and \emph{leg}. From each of the 14 ($= \binom{4}{1} + \binom{4}{2} + \binom{4}{3}$) non-empty proper subsets of labels, we randomly sample a collection of 200 labeled parts. Our hypothesis is that matrix rank should make it easy to distinguish between collections with few distinct labels, and collections with many distinct labels. Figure~\ref{fig:partFeatureDistance}~(right) plots the number of distinct labels in the part collection, vs increasing rank estimates. As we can see, all part collections with a single label have a lower score than those with two labels, which in turn are all lower than those with 3 labels. In contrast, a naive variance metric such as mean squared error, as shown in Figure~\ref{fig:partFeatureDistance}~(left), cannot correctly discriminate between part collections with 2 and 3 labels. We conclude that our rank-based metric accurately reflects consistency of a part collection.

\vspace{-8pt}

\paragraph{Control, adaptivity, and generalization.}
AdaCoSeg is not strongly supervised with consistently segmented and labeled training data, unlike most prior deep networks for shape segmentation. Instead, the weakly-supervised part prior allows a fair amount of input-dependent flexibility in what the actual co-segmentation looks like.

\begin{figure}[t!]
	\includegraphics[width=\linewidth]{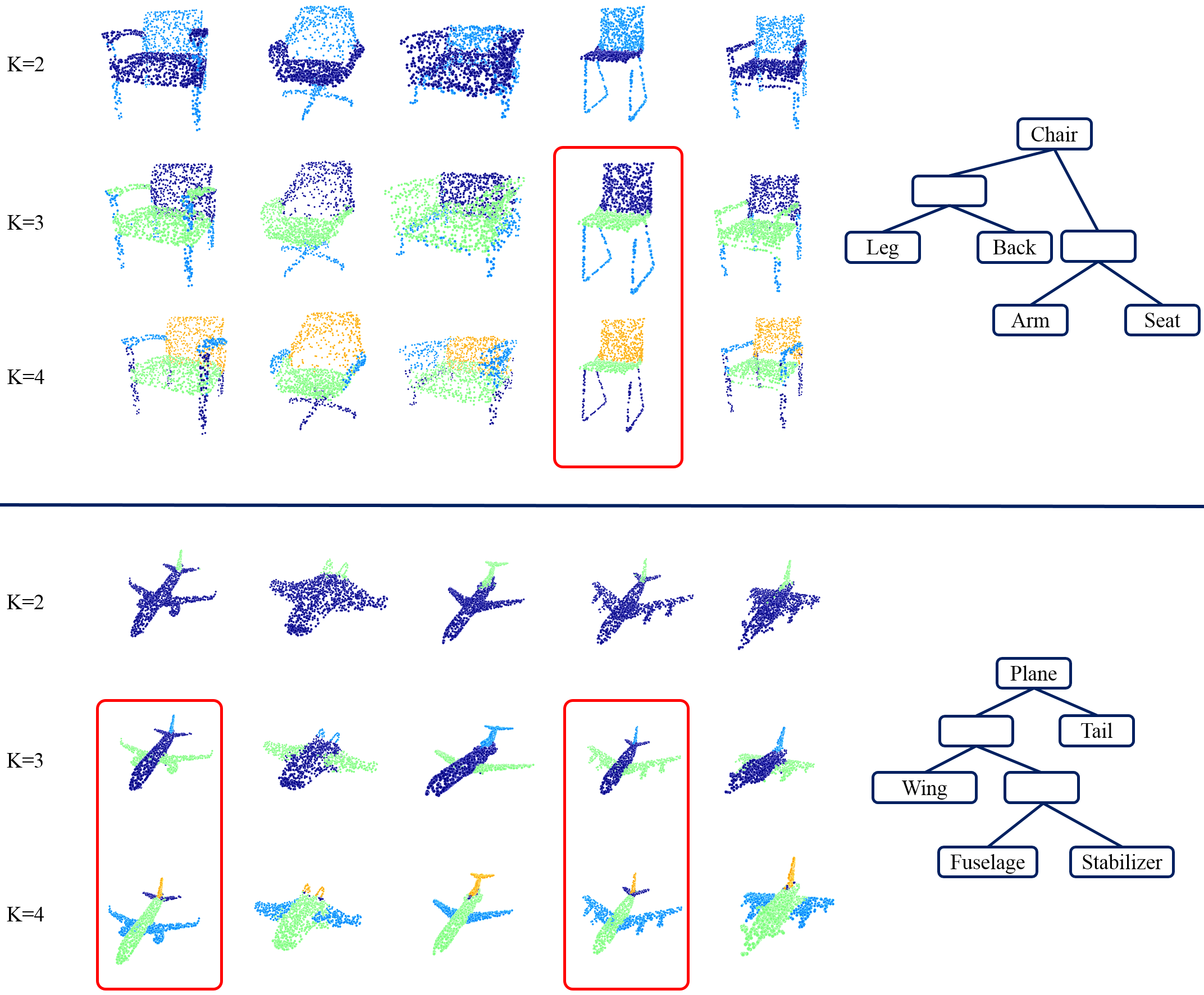}
	\caption{Coarse-to-fine co-segmentations of the same input shapes, generated by setting \mbox{$K = 2, 3, 4$}. The actual part count discovered per shape is adaptively selected and need not be exactly $K$, as shown in the examples bounded in red.}
	\label{fig:hierarchy}
\end{figure}

First, we can generate test set segmentations with different granularities, controlled by the cardinality bound $K$. Figure~\ref{fig:hierarchy} shows co-segmentation of the same shapes for different values of $K$. In these examples, our method fortuitously produces coarse-to-fine part hierarchies. However, this nesting structure is not guaranteed by the method, and we leave this as future work.

Further, even for a fixed $K$, different test shape collections can induce different co-segmentations. Figure~\ref{fig:teaser} shows co-segmentations of two different chair collections, both with \mbox{$K = 4$}. The collection on the left has several chairs with arms: hence, the optimization detects arms as one of the prominent parts and groups all chair legs into a single segment. The other collection has no arms, hence the four part types are assigned to back, seat, front, and back legs.

%\begin{figure}[t!]
%	\includegraphics[width=\linewidth]{figures/differentCoSegmentation.png}
%	\vspace{-7mm}
%	\caption{Co-segmentation results with the same $K=4$ for different sets of chairs, with or without arm. Our method can find the best segmentation granularity to fit the input $K$.}
%	\label{fig:differentcoseg}
%	\vspace{-2mm}
%\end{figure}

%We present two further experiments to show that our part prior network, trained on an unannotated offline database, is only a weak regularizer for co-segmentation. First, we show that reasonable co-segmentation results are obtained for a single category even when the offline training set includes shapes from multiple categories. Figure~\ref{fig:crosscategory}~(left) shows chairs co-segmented with the part prior network trained on both chairs and tables. Second, we show that weak denoising priors trained on one category can guide co-segmentation of another category. Figure~\ref{fig:crosscategory}~(right) shows tables co-segmented with the part prior network trained only on chairs.

%\begin{figure}[!t]
%	\includegraphics[width=\linewidth]{figures/crosscategory_new.png}
%	\caption{Some results of cross-category training for co-segmentation. \emph{Left}: Chairs co-segmented with a weak part prior trained on both chairs and tables. \emph{Right}: Tables co-segmented with a weak prior trained only on chairs.}
%	\label{fig:crosscategory}
%\end{figure}

\begin{figure*}[!t]
\centering
	\includegraphics[width=0.99\linewidth]{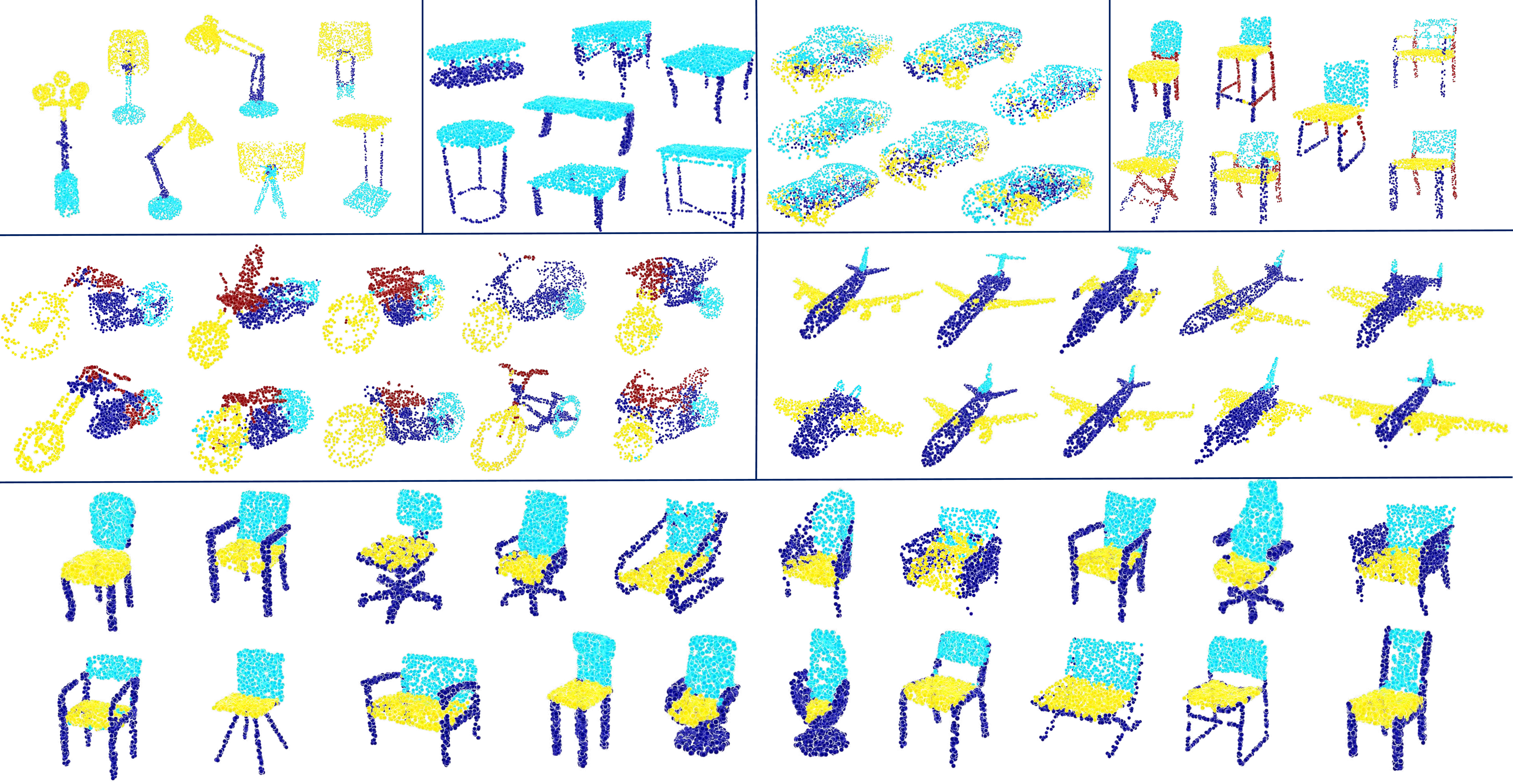}
	\caption{A gallery of co-segmentation results obtained by AdaCoSeg, for all the six object categories from the ComplementMe dataset. The input sets vary in size from 7 to 10. More results can be found in the supplementary material.}
	\label{fig:gallery}
\end{figure*}

\vspace{-8pt}

\paragraph{Quantitative evaluation.}
Since AdaCoSeg produces segmentations with varying granularity, it is difficult to compare its results to a fixed ground truth segmentation, e.g.,~\cite{wang2012active}. We adopt the following strategy. First, we set $K$ to be the total number of ground truth labels for a shape category. Second, after segmentation, we manually map our abstract labels $\{ L_1, L_2, \dots, L_K \}$ to the semantic labels (\emph{arm}, \emph{back}, \emph{wing} etc) present in the ground truth, using visual inspection of a few example shapes (this step could be automated, but it would not affect the overall argument). Now we can apply the standard Rand Index metric~\cite{chen2009benchmark} for segmentation accuracy:
\[
RI = 1 - \binom{2}{N}^{-1} \sum_{i < j}(C_{ij} P_{ij} + (1 - C_{ij}) (1 - P_{ij}))
\]
where $i,j$ are different points of the input point cloud. \mbox{$C_{ij} = 1$} iff $i$ and $j$ have the same predicted label, and \mbox{$P_{ij} = 1$} iff they have the same ground truth label. A lower Rand Index implies a better match with the ground truth. Note that the main advantage of RI over IOU is that it computes segmentation overlap without needing segment correspondence. This makes it particularly suited for evaluating co-segmentation where the focus is on segmentation consistency without knowing part labeling or correspondence.

In Table~\ref{tab:comparison}, we compare the Rand Index scores of our method vs prior work~\cite{sidi2011unsupervised,hu2012co,shu2016unsup}. Since our method trains category-specific weak priors by default, we evaluate on those categories of COSEG that are also present in the ComplementMe component dataset. Our method works natively with point clouds, whereas the three prior methods all have access to the original mesh data. Even so, we demonstrate the greatest overall accuracy (lowest RI).

\begin{figure}[t!]
	\includegraphics[width=0.99\linewidth]{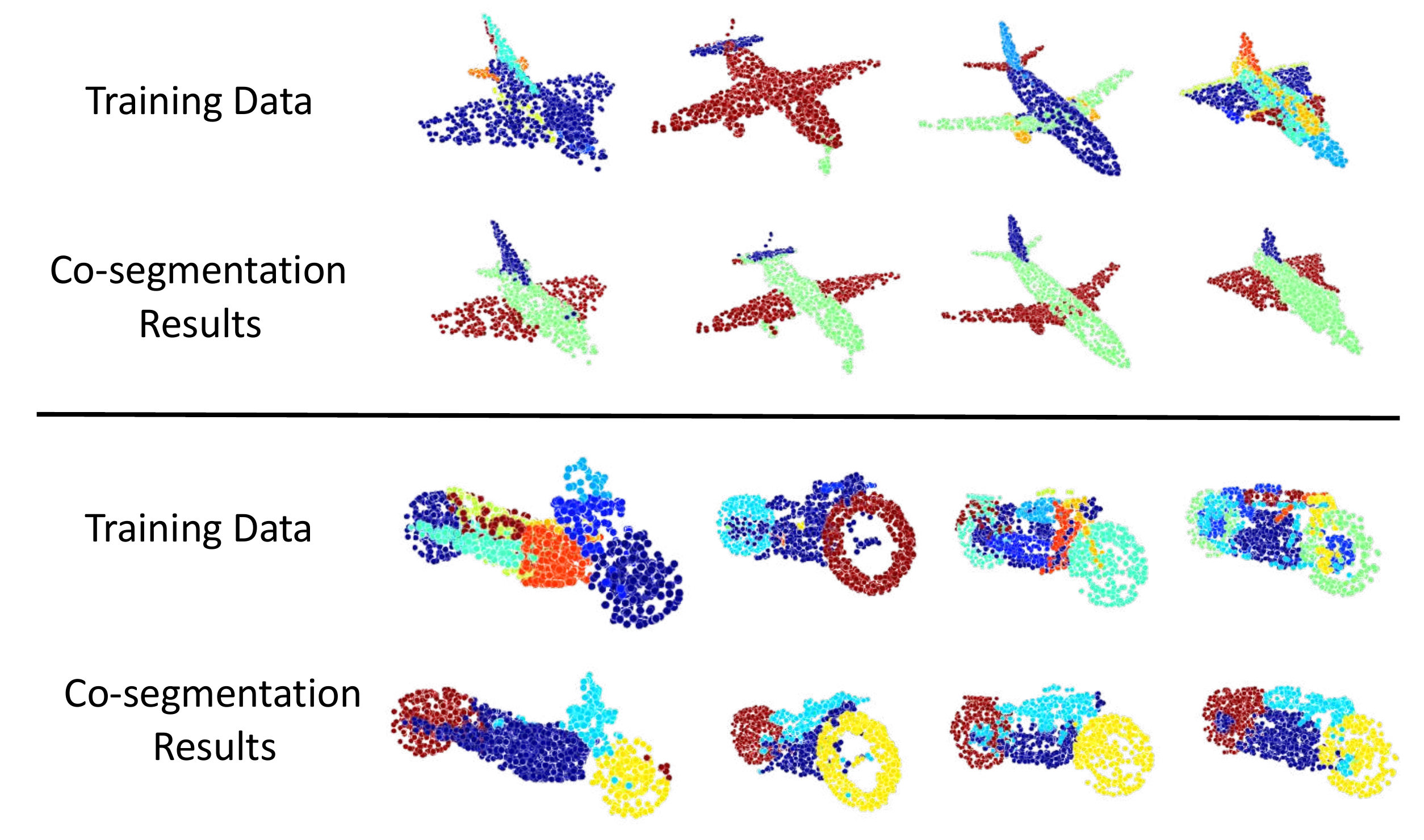}
	\caption{Co-segmentation results obtained by AdaCoSeg when using inconsistent training data. First and third rows show segmentations from the training data. Second and fourth rows show the co-segmentation results obtained by our network.}
	\label{fig:improveTraining}
\end{figure}

To demonstrate that AdaCoSeg does not rely on the initial training segmentations for the part prior network, we present a quantitative consistency evaluation between the initial segmentations and our co-segmentation results on a subset of our training data; the ground truth of this evaluation is labeled by experts. Table~\ref{tab:ImproveTraining} shows that AdaCoSeg can even improve the segmentation quality of its own training data. Figure~\ref{fig:improveTraining} demonstrates a significant improvement by our co-segmentation over the noisy training data. More results can be found in supplemental material.

\vspace{-8pt}

\begin{figure}[!t]
\centering
	\includegraphics[width=0.92\linewidth]{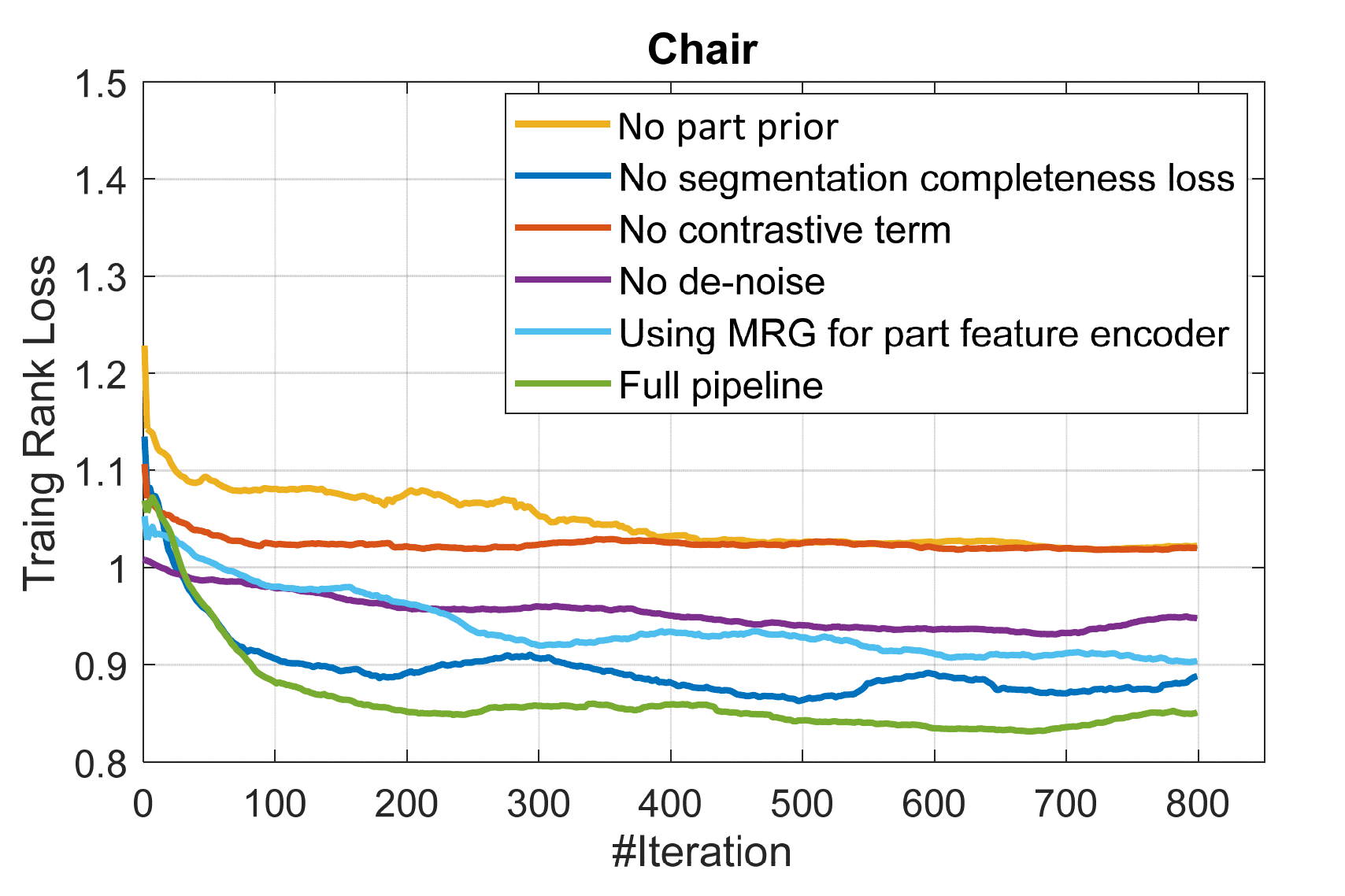}
	\caption{Training rank loss for ablation study on significant features. See supplemental material for more evaluation. }%\emph{Top:} Chair; \emph{Bottom:} Plane.}
	\label{fig:ablation}
\end{figure}

\paragraph{Ablation study.}
We explore the effect of our design choices via several ablation studies and show some results in Figure~\ref{fig:ablation}. These design choices include:
%\vspace{-8pt}
\begin{itemize}
\item {\em No part prior}: Remove the part prior network and connect the $K$-way classifier to point feature encoder. % directly.
\vspace{-3pt}  \item {\em No de-noise}: No random noise is added when training of our part prior network.
\vspace{-3pt}	\item {\em No segmentation completeness loss}: Optimize AdaCoSeg by using only the group consistency loss.
\vspace{-3pt}	\item {\em No contrastive term in group consistency loss}: Only keep the second term in our loss function.
\vspace{-3pt}	\item {\em MSG vs. MRG for part feature encoder}: Using MRG instead of MSG for encoding each shape part.
\end{itemize}

We found that the loss cannot decrease significantly without the part prior module and the contrastive term during training. Refer to the supplemental material for visual segmentation results without the part prior. Further, the denoising is also important for training our co-segmentation network. Finally, we found that the MSG feature for the part encoder, which focuses more on local than global contexts, can achieve better performance over MRG in our task.

\vspace{-8pt}

\begin{figure}[!t]
\centering
	\includegraphics[width=0.95\linewidth]{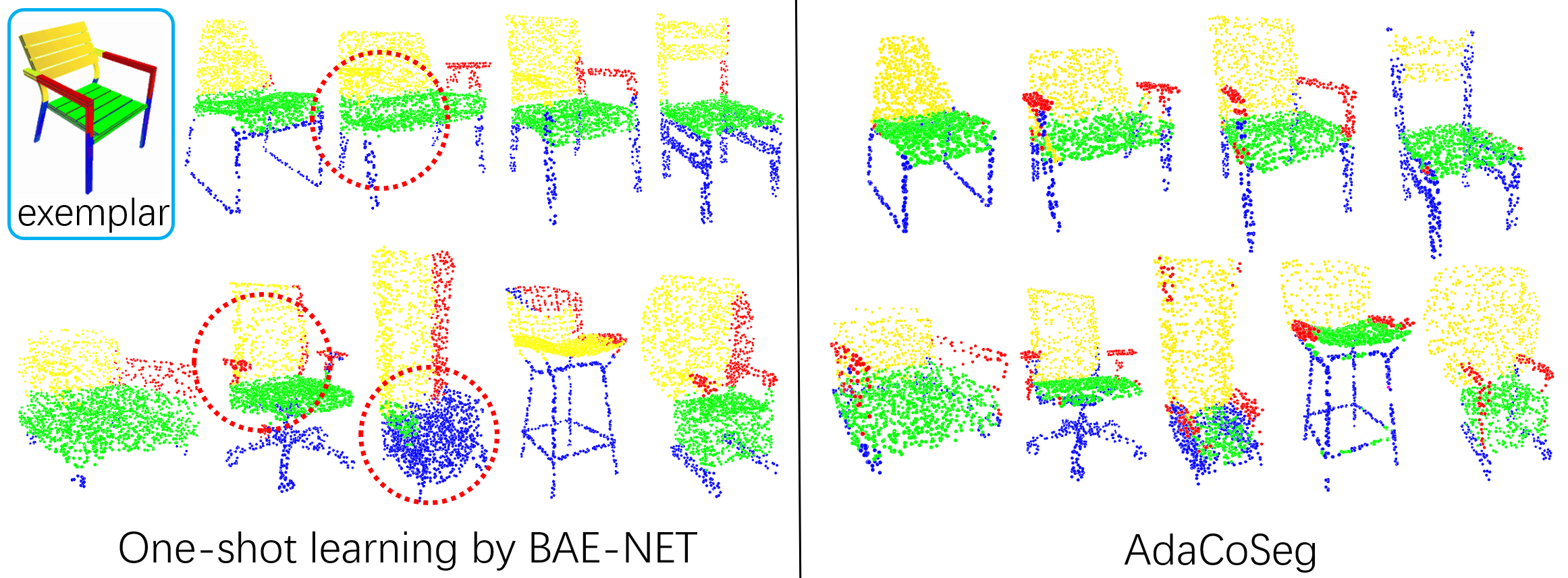}
	\caption{Comparing AdaCoSeg with BAE-NET on a small test set. AdaCoSeg, without needing any exemplars, leads to improved accuracy over BAE-NET with one exemplar.}
	\label{fig:comp-bae}
\end{figure}

\paragraph{Comparison to BAE-NET.}
\hao{Figure~\ref{fig:comp-bae} visually compares AdaCoSeg with one-shot learning of BAE-NET~\cite{chen2019baenet} using one perfect exemplar, on a small test set of 9 chairs; more comparison results can be found in the supplementary material. Both methods can be regarded as weakly supervised but with different supervision strategies. Our experiments show that with explicit optimization adapted to input sets, using the group consistency loss, AdaCoSeg generally outperforms BAE-NET over small test sets and in the presence of strong part discrepancies.}
%The part simplicity criterion followed by BAE-NET tends to produce coarser co-segmentations.}

%Please note that, the problem configuration of our method and BAE-NET are different while our final goal are the same. The co-segmentation in BAE-NET depends on a given segmented exemplar, which means the training framework of BAE-NET is one-shot supervised learning. The co-segmentation of our method depends on a trained part prior which can be noisy and inconsistent, it means that our training framework is weakly supervised. As the results shown in , we can find that even BAE-NET is given a perfect segmentation example, our method still has a better co-segmentation performance.}

\begin{table}[!t]
	\centering
	\begin{tabular}{c|c|c|c|c}
		Category & AdaCoSeg & Shu & Hu & Sidi \\
		\hline
		\hline
		Chair & {\bf 0.055} & $0.076$ & $0.121$ & $0.135$\\
		Lamp & {\bf 0.059} & $0.069$ & $0.103$ & $0.092$ \\
		Vase & $0.189$ & $0.198$ & $0.230$ & {\bf 0.102}\\
		Guitar & {\bf 0.032} & $0.041$ & $0.037$ & $0.081$
	\end{tabular}
	\vspace{2mm}
	\caption{Rand Index scores for AdaCoSeg vs.~prior works. \cy{With the exception of the vases, AdaCoSeg performs the best. The hand-crafted features from Sidi et al.~\cite{sidi2011unsupervised} prove to be best suited to the vase category.}}
	\label{tab:comparison}
\end{table}

\begin{table}[!t]
	\centering
	\begin{tabular}{c|c|c|c|c|c|c}
		 & {\small Chair} & {\small Table} & {\small Bicycle} & {\small Lamp} & {\small Car} & {\small Plane} \\
		\hline
		\hline
		{\small GT} & {\small 0.21} & {\small 0.27} & $0.31$ & {\small 0.18} & {\small 0.38} & {\small 0.24} \\
		{\small Ours} & {\small\bf 0.09} & {\small\bf 0.14} & {\small\bf 0.22} & {\small\bf 0.16} & {\small\bf 0.27} & {\small\bf 0.13}
	\end{tabular}
	\vspace{2mm}
	\caption{Rand Index score comparison between segmentations in training data (GT) and AdaCoSeg results. AdaCoSeg improves consistency even in its own training data. Visual results can be found in supplemental material.}
	\label{tab:ImproveTraining}
\end{table}

%\begin{table*}[!t]
%	\centering
%	\begin{tabular}{c|c|c|c|c|c|c}
%		{\bf Category} & Chair & Table & Bicycle & Lamp & Car & Airplane\\
%		\hline
%		\hline
%		Training set & $0.21$ & $0.27$ & $0.31$ & $0.18$ & $0.38$ & $0.24$ \\
%		Testing result & $0.09$ & $0.14$ & $0.22$ & $0.16$ & $0.27$ & $0.13$
%	\end{tabular}
%	\vspace{2mm}
%	\caption{Rand Index score comparison between segmentation in training data and the co-segmentation results given by our network. More related visual results can be found in supplemental material.}
%	\label{tab:ImproveTraining}
%\end{table*}

	\section{Conclusion, limitation, and future work}

We present AdaCoSeg, an adaptive deep learning framework for shape co-segmentation. A novel feature of our method is that \hao{beyond offline
training by the part prior netowk, the online co-segmentation network is adaptive to the input set of
shapes,} producing a consistent co-segmentation by iteratively minimizing a group consistency loss via backpropagation over a deep
network. Experiments demonstrate robustness of AdaCoSeg to large degrees 
of geometric and structural variations in the input sets, which is superior to state of the art.

No ground-truth consistent co-segmentations are needed to train AdaCoSeg. The offline and online stages are trained on 
different datasets, and for different tasks. The only supervision is at the first stage, to denoise part proposals on an individual shape 
basis, where the training can be carried out using existing datasets composed of inconsistent segmentations, e.g.,~\cite{sung2017comp}. 
The second optimizes a consistent segmentation on a specific test set, with the part prior as a regularizer. Our two-stage pipeline conserves 
computation by training the weak prior only once and reusing it across different co-segmentation tasks.
%

%, as well as co-segmentations at varying degrees of granularity.

\hao{
We reiterate that our online co-segmentation network does {\em not\/} generalize to new inputs, which is by design: the network weights are 
derived to minimize the loss function for the current input set and they are recomputed for each new set. 
%
%When two input sets are sufficiently similar, the learned weights for one set could serve as a good starting point for the other set to save optimization time.}
%
% Results reveal that this is true when the sets are sufficiently similar.} When this is not the case, the network 
%may be stuck in a local minima and the resulting co-segmentations are not meaningful. %; this is an expected outcome of gradient decent.
%
%
Also, AdaCoSeg is not trained end-to-end. While an end-to-end deep co-segmentation network is desirable,
the challenges of developing such networks for an unsupervised problem are well known~\cite{DBLP:journals/corr/abs-1811-12359}.}
%
%That being said, our part prior network does not perform a pre-segmentation, with the following online stage refining it. The two stages 
%in CoSegNet are trained on different datasets, and for different tasks. The first trains a general part prior on a large unlabeled dataset, 
%which can be reused across different co-segmentation tasks. The second optimizes a consistent segmentation on a specific test set, with 
%the part prior as a regularizer. Our two-stage pipeline conserves computation by training the weak prior only once and reusing it.}
%
Another limitation is that our part prior network is not trained across different object categories. \hao{This would have been ideal, but 
per-category training is typical for most existing segmentation models~\cite{huang2018recurrent,klokov2017escape,riegler2017octnet,wang2017cnn}.} Our current
network appears capable of handling some intra-category variations,
%even cross-category training when the categories share much similarity in their part characteristics (e.g., tables and chairs),
but learning parts and their feature descriptions with all categories mixed together is significantly more challenging.

In future work, we plan to extend our weakly supervised learning framework for cross-category part learning. We would also like to explore
co-segmentation via {\em online learning\/}, which represents a family of machine learning
algorithms that learn to update models incrementally from sequentially input data streams~\cite{shalev2007online,jin2010online}.
In contrast, our current co-segmentation network does not really learn a generalizable model, and the learned network weights cannot be continuously
updated as new shapes come in. An online learned model for unsupervised co-segmentation may need to
create and maintain multiple segmentation templates.

\section*{Acknowledgement}
We thank all the anonymous reviewers for their valuable comments and suggestions. This work was supported in part by an NSERC grant (No.~611370), NSFC (61572507, 61532003, 61622212), NUDT Research Grants (No.~ZK19-30), National Key Research and Development Program of China (No.~2018AAA0102200), NSF grants CHS-1528025 and IIS-1763268, a Vannevar Bush Faculty Fellowship, a grant from the Dassault Foundation, a Natural Science Foundation grant for Distinguished Young Scientists (2017JJ1002) from the Hunan Province, and gift funds from Adobe.

	\bibliographystyle{ieee_fullname}
	\bibliography{ms}

\begin{thebibliography}{10}\itemsep=-1pt

\bibitem{DBLP:journals/corr/BadrinarayananK15}
Vijay Badrinarayanan, Alex Kendall, and Roberto Cipolla.
\newblock {SegNet}: A deep convolutional encoder-decoder architecture for image
  segmentation.
\newblock {\em TPAMI}, 39(12):2481--2495, 2017.

\bibitem{chang2015shapenet}
Angel~X Chang, Thomas Funkhouser, Leonidas Guibas, Pat Hanrahan, Qixing Huang,
  Zimo Li, Silvio Savarese, Manolis Savva, Shuran Song, Hao Su, et~al.
\newblock {ShapeNet}: An information-rich {3D} model repository.
\newblock {\em arXiv preprint arXiv:1512.03012}, 2015.

\bibitem{DBLP:journals/corr/ChenPK0Y16}
Liang{-}Chieh Chen, George Papandreou, Iasonas Kokkinos, Kevin Murphy, and
  Alan~L. Yuille.
\newblock {DeepLab}: Semantic image segmentation with deep convolutional nets,
  atrous convolution, and fully connected {CRFs}.
\newblock {\em CoRR}, abs/1606.00915, 2016.

\bibitem{chen2009benchmark}
Xiaobai Chen, Aleksey Golovinskiy, and Thomas Funkhouser.
\newblock A benchmark for {3D} mesh segmentation.
\newblock In {\em Trans. Graph.}, volume~28, 2009.

\bibitem{chen2019baenet}
Zhiqin Chen, Kangxue Yin, Matt Fisher, Siddhartha Chaudhuri, and Hao Zhang.
\newblock {BAE-NET}: Branched autoencoder for shape co-segmentation.
\newblock In {\em ICCV}, 2019.

\bibitem{golovinskiy2009consistent}
Aleksey Golovinskiy and Thomas Funkhouser.
\newblock Consistent segmentation of {3D} models.
\newblock {\em Computers \& Graphics}, 33(3):262--269, 2009.

\bibitem{hu2012co}
Ruizhen Hu, Lubin Fan, and Ligang Liu.
\newblock Co-segmentation of {3D} shapes via subspace clustering.
\newblock {\em Computer Graphics Forum}, 31(5):1703--1713, 2012.

\bibitem{huang2011joint}
Qixing Huang, Vladlen Koltun, and Leonidas Guibas.
\newblock Joint shape segmentation with linear programming.
\newblock {\em Trans. Graph.}, 30(6), 2011.

\bibitem{huang2018recurrent}
Qiangui Huang, Weiyue Wang, and Ulrich Neumann.
\newblock Recurrent slice networks for {3D} segmentation of point clouds.
\newblock In {\em CVPR}, 2018.

\bibitem{jin2010online}
Rong Jin, Steven~CH Hoi, and Tianbao Yang.
\newblock Online multiple kernel learning: Algorithms and mistake bounds.
\newblock In {\em Int'l Conf. on Algorithmic Learning Theory}, 2010.

\bibitem{kalogerakis20173d}
Evangelos Kalogerakis, Melinos Averkiou, Subhransu Maji, and Siddhartha
  Chaudhuri.
\newblock {3D} shape segmentation with projective convolutional networks.
\newblock In {\em CVPR}, 2017.

\bibitem{kalogerakis2010crf}
Evangelos Kalogerakis, Aaron Hertzmann, and Karan Singh.
\newblock Learning {3D} mesh segmentation and labeling.
\newblock {\em Trans. Graph. (SIGGRAPH)}, 29(3), 2010.

\bibitem{kar2014nondecomp}
Purushottam Kar, Harikrishna Narasimhan, and Prateek Jain.
\newblock Online and stochastic gradient methods for non-decomposable loss
  functions.
\newblock In {\em NeurIPS}, 2014.

\bibitem{kingma2015adam}
Diederik~P. Kingma and Jimmy Ba.
\newblock Adam: A method for stochastic optimization.
\newblock In {\em ICLR}, 2015.

\bibitem{klokov2017escape}
Roman Klokov and Victor Lempitsky.
\newblock Escape from cells: Deep kd-networks for the recognition of {3D} point
  cloud models.
\newblock In {\em ICCV}, 2017.

\bibitem{li2018deep}
Weihao Li, Omid~Hosseini Jafari, and Carsten Rother.
\newblock Deep object co-segmentation.
\newblock In {\em ACCV}, 2018.

\bibitem{DBLP:journals/corr/abs-1811-12359}
Francesco Locatello, Stefan Bauer, Mario Lucic, Sylvain Gelly, Bernhard
  Sch{\"{o}}lkopf, and Olivier Bachem.
\newblock Challenging common assumptions in the unsupervised learning of
  disentangled representations.
\newblock In {\em ICML}, 2019.

\bibitem{mitra2013structure}
Niloy Mitra, Michael Wand, Hao~Richard Zhang, Daniel Cohen-Or, Vladimir Kim,
  and Qi-Xing Huang.
\newblock Structure-aware shape processing.
\newblock In {\em SIGGRAPH Asia 2013 Courses}, 2013.

\bibitem{muralikrishnan2018tags2parts}
Sanjeev Muralikrishnan, Vladimir~G Kim, and Siddhartha Chaudhuri.
\newblock {Tags2Parts}: Discovering semantic regions from shape tags.
\newblock In {\em CVPR}, 2018.

\bibitem{qi2017pnpp}
Charles~Ruizhongtai Qi, Li Yi, Hao Su, and Leonidas~J Guibas.
\newblock {PointNet++}: Deep hierarchical feature learning on point sets in a
  metric space.
\newblock In {\em NeurIPS}, 2017.

\bibitem{riegler2017octnet}
Gernot Riegler, Ali~Osman Ulusoy, and Andreas Geiger.
\newblock {OctNet}: Learning deep {3D} representations at high resolutions.
\newblock In {\em CVPR}, 2017.

\bibitem{shalev2007online}
Shai Shalev-Shwartz and Yoram Singer.
\newblock Online learning: Theory, algorithms, and applications.
\newblock 2007.

\bibitem{shu2016unsup}
Zhenyu Shu, Chengwu Qi, Shiqing Xin, Chao Hu, Li Wang, Yu Zhang, and Ligang
  Liu.
\newblock Unsupervised {3D} shape segmentation and co-segmentation via deep
  learning.
\newblock {\em Computer Aided Geometric Design}, 43:39--52, 2016.

\bibitem{sidi2011unsupervised}
Oana Sidi, Oliver van Kaick, Yanir Kleiman, Hao Zhang, and Daniel Cohen-Or.
\newblock Unsupervised co-segmentation of a set of shapes via descriptor-space
  spectral clustering.
\newblock {\em Trans. Graph. (SIGGRAPH Asia)}, 30(6), 2011.

\bibitem{sung2017comp}
Minhyuk Sung, Hao Su, Vladimir~G. Kim, Siddhartha Chaudhuri, and Leonidas
  Guibas.
\newblock {ComplementMe}: Weakly-supervised component suggestions for {3D}
  modeling.
\newblock {\em Trans. Graph. (SIGGRAPH Asia)}, 2017.

\bibitem{sung2018deep}
Minhyuk Sung, Hao Su, Ronald Yu, and Leonidas Guibas.
\newblock Deep functional dictionaries: Learning consistent semantic structures
  on {3D} models from functions.
\newblock In {\em NeurIPS}, 2018.

\bibitem{vicente2011object}
Sara Vicente, Carsten Rother, and Vladimir Kolmogorov.
\newblock Object cosegmentation.
\newblock In {\em CVPR}, 2011.

\bibitem{wang20183d}
Pengyu Wang, Yuan Gan, Panpan Shui, Fenggen Yu, Yan Zhang, Songle Chen, and
  Zhengxing Sun.
\newblock {3D} shape segmentation via shape fully convolutional networks.
\newblock {\em Computers \& Graphics}, 70:128--139, 2018.

\bibitem{wang2017cnn}
Peng-Shuai Wang, Yang Liu, Yu-Xiao Guo, Chun-Yu Sun, and Xin Tong.
\newblock {O-CNN}: Octree-based convolutional neural networks for {3D} shape
  analysis.
\newblock {\em ACM Transactions on Graphics}, 36(4), 2017.

\bibitem{wang2012active}
Yunhai Wang, Shmulik Asafi, Oliver Van~Kaick, Hao Zhang, Daniel Cohen-Or, and
  Baoquan Chen.
\newblock Active co-analysis of a set of shapes.
\newblock {\em Trans. Graph. (SIGGRAPH Asia)}, 31(6), 2012.

\bibitem{xu2010style}
Kai Xu, Honghua Li, Hao Zhang, Daniel Cohen-Or, Yueshan Xiong, and Zhi-Quan
  Cheng.
\newblock Style-content separation by anisotropic part scales.
\newblock {\em Trans. Graph. (SIGGRAPH Asia)}, 29(6), 2010.

\bibitem{yi2016scalable}
Li Yi, Vladimir~G Kim, Duygu Ceylan, I Shen, Mengyan Yan, Hao Su, Cewu Lu,
  Qixing Huang, Alla Sheffer, Leonidas Guibas, et~al.
\newblock A scalable active framework for region annotation in {3D} shape
  collections.
\newblock {\em Trans. Graph. (SIGGRAPH Asia)}, 35(6), 2016.

\bibitem{yi2017syncspeccnn}
Li Yi, Hao Su, Xingwen Guo, and Leonidas~J Guibas.
\newblock {SyncSpecCNN}: Synchronized spectral {CNN} for {3D} shape
  segmentation.
\newblock In {\em CVPR}, 2017.

\bibitem{yi2018faces}
Renjiao Yi, Chenyang Zhu, Ping Tan, and Stephen Lin.
\newblock Faces as lighting probes via unsupervised deep highlight extraction.
\newblock In {\em ECCV}, 2018.

\end{thebibliography}

\end{document}